\documentclass[conference]{IEEEtran}
\IEEEoverridecommandlockouts
\usepackage{cite}
\usepackage{xcolor}
\def\BibTeX{{\rm B\kern-.05em{\sc i\kern-.025em b}\kern-.08em
    T\kern-.1667em\lower.7ex\hbox{E}\kern-.125emX}}
    
\usepackage{float}
\usepackage{hyperref}
\usepackage{comment}
\usepackage{gensymb} 
\usepackage[flushleft]{threeparttable}
\usepackage{multirow}
\usepackage{amsmath}
\usepackage{amssymb}
\usepackage{bm}
\usepackage{graphicx}
\usepackage{subfig}
\usepackage[linesnumbered,ruled,vlined]{algorithm2e}
\usepackage{algorithmic}
\usepackage[utf8]{inputenc}
\usepackage[T1]{fontenc}
\usepackage{textcomp}
\usepackage{nomencl}
\usepackage{rotating}
\usepackage{array}
\usepackage{mathtools}
\usepackage{diagbox}
\usepackage{upgreek}
\usepackage{url}
\usepackage{footnote} 
\usepackage{slashbox} 
\usepackage{bm} 
\usepackage[thinc]{esdiff}
\usepackage{lipsum}
\usepackage{cuted}

\title{\LARGE \bf
Noise Tolerant Identification and Tuning Approach Using Deep Neural Networks For Visual Servoing Applications
\thanks{This work was supported by Khalifa University grants CIRA-2020-082 and RC1-2018-KUCARS. Oussama Abdul Hay is the corresponding author (\emph{email: oussama.hay@ku.ac.ae}).

O. Abdul Hay, M. Chehadeh, A. Ayyad, M. Wahbah, M. Humais, I. Boiko, and Y. Zweiri are with the Center for Autonomous Robotic Systems, Khalifa University, Abu Dhabi, United Arab Emirates. Also, I. Boiko is with the Department of Electrical Engineering and Computer Science, and Y. Zweiri with the Department of Aerospace Engineering, both at Khalifa University, Abu Dhabi, United Arab Emirates.}
}

\author{
        Oussama~Abdul~Hay\href{https://orcid.org/0000-0001-8299-6021}{\includegraphics[scale=0.75]{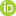}},
        Mohamad~Chehadeh\href{https://orcid.org/0000-0002-9430-3349}{\includegraphics[scale=0.75]{Figures/orcid.png}},~\IEEEmembership{Member,~IEEE,}
        Abdulla~Ayyad\href{https://orcid.org/0000-0002-3006-2320}{\includegraphics[scale=0.75]{Figures/orcid.png}},~\IEEEmembership{Member,~IEEE,}\\
        Mohamad~Wahbah\href{https://orcid.org/0000-0003-1647-8546}{\includegraphics[scale=0.75]{Figures/orcid.png}},
        Muhammad~Humais\href{https://orcid.org/0000-0001-6237-6394}{\includegraphics[scale=0.75]{Figures/orcid.png}},
        Igor~Boiko\href{https://orcid.org/0000-0003-4978-614X}{\includegraphics[scale=0.75]{Figures/orcid.png}},~\IEEEmembership{Senior~Member,~IEEE,},
        Lakmal~Seneviratne\href{https://orcid.org/0000-0001-6405-8402}{\includegraphics[scale=0.75]{Figures/orcid.png}}\\
        and~Yahya~Zweiri\href{https://orcid.org/0000-0003-4331-7254}{\includegraphics[scale=0.75]{Figures/orcid.png}},~\IEEEmembership{Member,~IEEE}

\thanks{*Oussama Abdul Hay and Mohamad Chehadeh contributed equally to this work.}

}

\begin{document}

\maketitle
\thispagestyle{empty}
\pagestyle{empty}

\begin{abstract}
Vision based control of Unmanned Aerial Vehicles (UAVs) has been adopted by a wide range of applications due to the availability of low-cost on-board sensors and computers. Tuning such systems to work properly requires extensive domain specific experience, which limits the growth of emerging applications. Moreover, obtaining performance limits of UAV based visual servoing is difficult due to the complexity of the models used. In this paper, we propose a novel noise tolerant approach for real-time identification and tuning of visual servoing systems, based on deep neural networks (DNN) classification of system response generated by the modified relay feedback test (MRFT). The proposed method, called DNN with noise protected MRFT (DNN-NP-MRFT), can be used with a multitude of vision sensors and estimation algorithms despite the high levels of sensor's noise. Response of DNN-NP-MRFT to noise perturbations is investigated and its effect on identification and tuning performance is analyzed. The proposed DNN-NP-MRFT is able to detect performance changes due to the use of high latency vision sensors, or due to the integration of inertial measurement unit (IMU) measurements in the UAV states estimation. Experimental identification closely matches simulation results, which can be used to explain system behaviour and predict the closed loop performance limits for a given hardware and software setup. We also demonstrate the ability of DNN-NP-MRFT tuned UAVs to reject external disturbances like wind, or human push and pull. Finally, we discuss the advantages of the proposed DNN-NP-MRFT visual servoing design approach compared with other approaches in literature.
\end{abstract}

\section{INTRODUCTION}
\subsection{Motivation}
A variety of Unmanned Aerial Vehicles (UAVs) applications require visual servoing tasks as part of their overall mission. Many examples can be found in the literature for such tasks like mission control \cite{Thomas2016}, infrastructure inspection \cite{Mathe2015}, and aerial manipulation \cite{BONYANKHAMSEH2018221}. From a broad perspective, visual servoing consists of three elements: a vision sensor that provides a projection of the surrounding physical world, vision and state estimation algorithms that are used for data interpretation, and control algorithms that drive system states to desired references. Recent work on visual servoing for UAVs \cite{lippiello2015hybrid,Thomas2016} addressed the need for higher order dynamic models that describe UAVs' motion in order to achieve better control performance. Still, these models are simplified in the sense that the dynamics of the vision sensors, and vision and state estimation algorithms are absent. Although the authors in \cite{lippiello2015hybrid,Thomas2016,thomas2017autonomous} presented system stability proofs, they did not present guarantees on system performance for a given set of controller parameters. Other researchers reported that addressing these additional dynamics of sensors and actuators are important to achieve optimal performance. For example, \cite{mueggler2014event} reported that the use of event cameras instead of vision cameras improved the dynamic performance of UAVs. Also, \cite{Chuang2019tracking} reported a significant change in the tracking performance of a moving visual target when the update rate of the visual detection is changed. The role of such parasitic dynamics in the tuning of high-performance systems is becoming widely acknowledged in the general robotics community. The effect of perception latency was studied in \cite{Falanga2019toofast} for high speed sense and avoid applications. The value of latency was estimated using camera parameters. A drawback of this approach is that it requires careful analysis of the impact of the vision-based mission, as well as the used hardware and software setup on latency characteristics. 
Controller tuning becomes less systematic and heavily dependent on human-expert knowledge as a result of omitting such important parasitic dynamics, All the visual servoing tasks presented in \cite{oliveira2013ground,borowczyk2017autonomous,zhang2018vision} used PID controllers with no systematic strategy for controller parameters tuning. Tuning such controllers is not a straightforward task given particularly since UAVs are underactuated where the interaction between the outer loops and the inner loops can be complex. In \cite{cheng2017autonomous} the authors attempted to reduce this complexity by designing a non-linear controller with a reduced number of tunable parameters, but servoing performance was poor. A main drawback of this approach is losing system linearity.

For the aforementioned reasons, in-flight automated controller tuning has recently gained increased interest in the robotics literature. The in-flight tuning method proposed in \cite{horla2021tune} was demonstrated for the tuning of altitude PD controllers. This method does not run in real-time and requires a few minutes for the controller parameters to converge. Another non real-time in-flight tuning approach applied to hex-rotor UAVs with tilting propellers was presented in \cite{horla2021optimal}.The controller trajectory tracking and regulation performance in \cite{horla2021tune,horla2021optimal} was not demonstrated. The work in \cite{Chehadeh2019} developed non-parametric tuning rules for near-optimal real-time in-flight tuning of attitude and altitude PD controllers based on the modified relay feedback test (MRFT). The accuracy of tuning in \cite{Chehadeh2019} was further improved by the introduction of deep neural networks and the modified relay feedback test (DNN-MRFT) identification and tuning approach in \cite{ayyad2020real}. The DNN-MRFT improved over the non-parametric MRFT based approaches \cite{Chehadeh2019,Boiko2013book} due to the use of DNN, which provides a mapping between MRFT excited oscillations and process parameters based on the oscillation shape, frequency, and bias of the unknown system. The DNN was solely trained based on simulation data. The DNN-MRFT approach was extended to underactuated lateral motion loops in \cite{ayyad2021multirotors} through a novel cascaded identification method. DNN-MRFT application to UAV visual servoing was not successful, due to the high levels of sensor noise, resulting in false switching of the MRFT as demonstrated in the attached supplementary material. MRFT noise mitigation algorithms, developed for loops widely present in the process industry and described in \cite{Boiko2013book}, fail at preventing false switching in UAV visual servoing. The research presented in this paper was motivated by the limitations of the current DNN-MRFT for UAV visual servoing tasks.

\subsection{Contributions}
The main contribution of this paper is the development of a novel noise tolerant identification and tuning method for UAVs visual servoing applications. The proposed approach is based on a deep neural network with noise protected modified relay feedback test (DNN-NP-MRFT) that mitigates the effects of sensor noise while maintaining the identification and tuning performance of DNN-MRFT. A novel switching algorithm for MRFT is introduced to avoid false switching caused by the high frequency noise. 

The switching algorithm restricts the value of the MRFT parameter \(\beta\) to \(\beta_{min}\) to generate limit cycles. We show that this modification does not alter the identification accuracy if \(\beta_{min}\) is less than the optimal test phase \(\beta^*\). Predictability of system performance is demonstrated by the match between simulated identified model responses and experimental responses. The effect of the use of different sensors, and sensor fusion structures on the performance of system dynamics is demonstrated using DNN-NP-MRFT as a unifying framework. The sensors used are a normal camera (a frame-based visible wavelength camera), an event based camera (also known as dynamic vision sensor or neuromorphic vision sensor), and a thermal camera. DNN-NP-MRFT is a lightweight algorithm where online data preprocessing and DNN prediction takes just a few milliseconds with a modern embedded processors, and is suitable for real-time identification and tuning. To the best of our knowledge, this is the first reporting in literature of a real-time identification and tuning method applied for UAVs that use on-board sensors for control. The proposed approach was verified experimentally where the UAV achieved near optimal step reference following performance, with real-time tuning and without requiring any additional manual tuning. The demonstrated optimality covers a wide range of frequency spectra of the closed loop system as it minimizes the integral square error (ISE) metric. The tuned system was able to withstand wind disturbances as high as 5 m/s, and external pull and release disturbances up to 10 N.

\subsection{Scope}
The proposed tuning method is applicable to a wide variety of UAV visual servoing tasks, including image based visual servoing (IBVS) and position based visual servoing (PBVS). Without loss of generality, the proposed approach is evaluated in this paper by commanding the UAV to lock and move relative to a target that is fixed to a planar vertical structure. In another test case, the target moves arbitrarily in discrete large steps and the UAV keeps following the target. Such step changes are not uncommon in visual servoing scenarios, e.g. recovery from occlusions or temporal out of field of view (FOV) state. When a Kalman Filter (KF) is used, the distance to target is assumed to be known, thus results with KF assumes a PBVS scenario and results without the KF resembles an IBVS scenario. This arrangement was inspired by Challenge 3 of Mohamed Bin Zayed International Robotics Challenge 2020 (MBZIRC2020) \cite{mbzirc2020} where firefighting UAVs were used to target fires both indoors and outdoors. Intuitively, the same techniques presented in this paper can be applied to other arrangements like servoing an object on the ground. Tuning controllers to follow trajectories can be considered as an extension to this work where optimality with regards to excitation sources other than a step could be used. 

An important advantage of the proposed DNN-NP-MRFT approach is its ease of applicability to new UAV designs, i.e. there are no parameters specific to a particular UAV configuration that needs to be chosen a priori, except for the noise rejection parameter \(\tau_{obs}\) which is specific to the sensor and UAV used. Choosing this parameter is still straightforward, and it is rarely changed for new UAV designs as will be shown in the discussion about DNN-NP-MRFT in Section \ref{sec:robust_mrft}. As a result, the existing advantages associated with the general applicability of DNN-MRFT to a wide variety of UAV designs are still retained with DNN-NP-MRFT approach.

\section{State Estimation}
The main system components of the visual servoing design suggested in this paper are outlined in Fig. \ref{fig:identification_structure}. The proposed system will operate in one of two phases: identification phase or control phase. In this section we discuss various aspects of state estimation, and in the next section we discuss modeling and identification through DNN-NP-MRFT.

\begin{figure}
    \centering
    \includegraphics[width=0.4\textwidth]{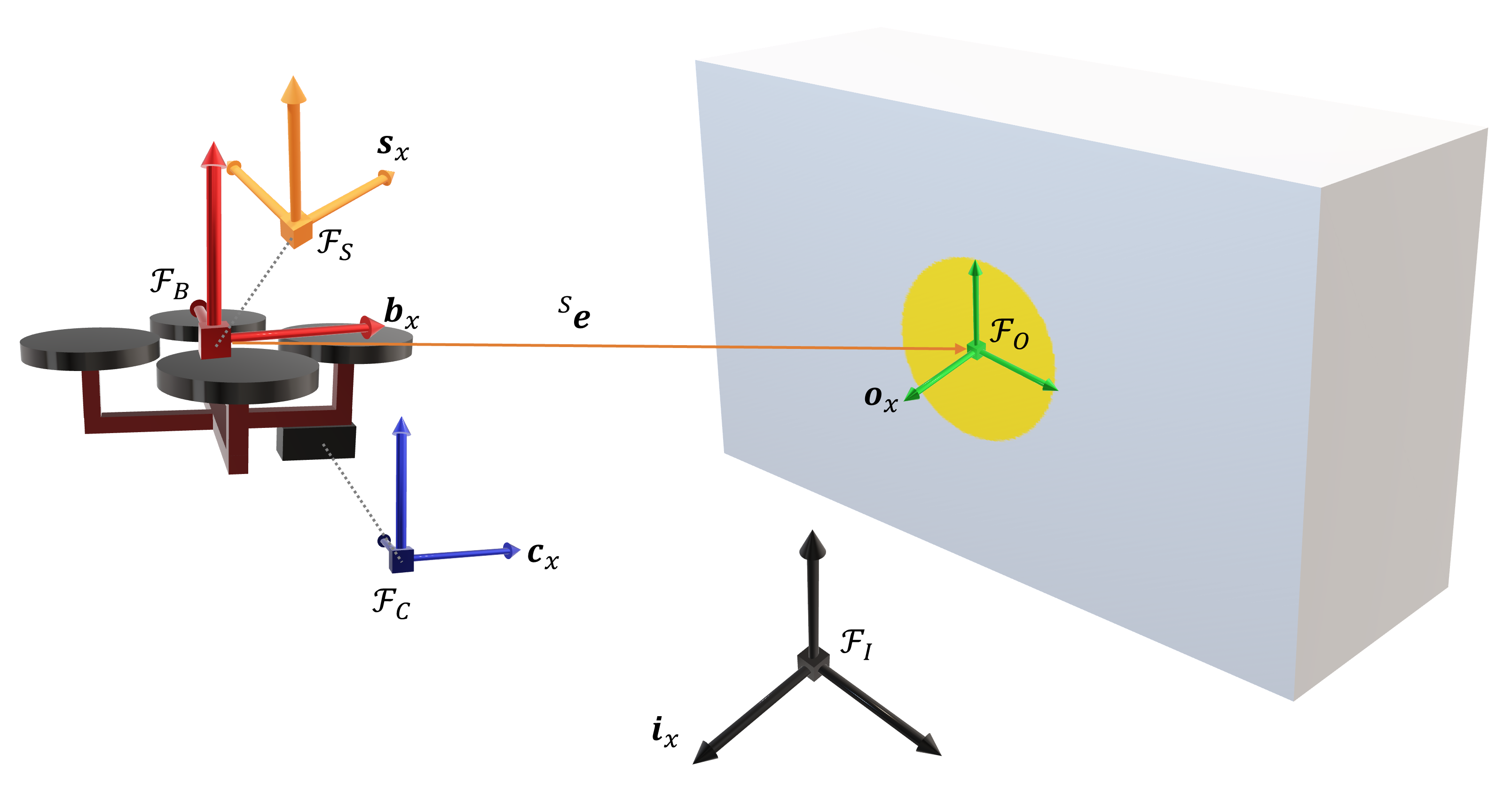}
    \caption{Illustration of the reference frames used in this paper. Note that the camera reference frame \(\mathcal{F}_C\) and servoing reference frame \(\mathcal{F}_S\) are offset for visual clarity. The axis \(\bm{s_x}\) is defined to be anti-aligned with the axis \(\bm{o_x}\).}
    \label{fig:reference_frame}
\end{figure}

In this work, we define the inertial frame \(\mathcal{F}_I\) to be earth-fixed global reference frame with \(\bm{i_z}\) axis pointing upwards. The body reference frame \(\mathcal{F}_B\) is attached to the multirotor UAV center of mass, with axis \(\bm{b_z}\) perpendicular to the propulsion plane, and pointing in the thrust direction. Another body attached reference frame is the camera reference frame \(\mathcal{F}_C\) has \(\bm{c_x}\) axis aligned with the focal vector of the camera and centered at the sensor center. For convenience, we define another reference frame \(\mathcal{F}_O\) that is attached to the tracked object with \(\bm{o_x}\) axis always perpendicular to the object plane. The last used reference frame is the servoing reference frame \(\mathcal{F}_S\) has its origin coinciding with \(\mathcal{F}_B\) origin, and with axis \(\bm{s_x}=-\bm{o_x}\). The servoing error is then defined as (note that all vectors are in \(\mathcal{F}_S\)):
\begin{equation}
\label{eq:error_function}
    {}^S\bm{e}={}^S\bm{r_c}-{}^S\bm{c}
\end{equation}
where \({}^S\bm{c}\), and \({}^S\bm{r_c}\) are the object center and desired reference respectively (all distance units are in meters). Note that the error is negated here, compared to the convention widely used in visual servoing literature \cite{chaumette2016visual}. Fig. \ref{fig:reference_frame} provides a graphical illustration of the utilized reference frames and the error vector.
All the reference frames are right-handed and a counter clockwise rotation is positive by definition. A pose transformation defined in \(SE(3)\) Lie group used to transform between reference frames is denoted by \({}^B_AT\), where \(B\) is the target reference frame and \(A\) is the source one, and where \(T\) is a 4 by 4 matrix with unity homogeneous coordinates.

\subsection{Object Detection and Position Estimation} \label{sec:obj_det}
Without loss of generality, we simplify the object detection task by assuming prior knowledge of the object shape. During visual servoing, the UAV position is referenced to the center of the detected object. We have used a screen projector to project shapes on the screen wall. With this setup, we can easily create different object shapes and program their motion.
For the normal and event cameras, the task was to track a circle. The detection pipeline included a blob detector to retrieve the circle's center \(\bm{o}=[o_x \; o_y]^T\) in pixel coordinates. For the event camera, we use a checkered circle that is rotating at a fixed speed to ensure continuous generation of events. Still, the generated events do not fully form a circle, and a connected set algorithm is used to form a circle. We used a rectangular thermal source to be detected by the thermal camera, where a simple threshold is applied to mask the thermal source. Changes to the detected shape has no effect on the followed methodology.

To recover relative position from detected object center, we use the pinhole camera model given by \cite{chaumette2016visual}:
\begin{align}
\begin{split}
    \begin{cases}
      p_y = \frac{c_y}{c_x} = \frac{o_x-c_u}{f\alpha}\\
      p_z = \frac{c_z}{c_x} = \frac{o_y-c_v}{f}
    \end{cases}
\end{split}
\end{align}
where \({}^C\bm{p}=[f\, p_y\, p_z]^T\) is the projected center on the image plane, \({}^C\bm{c}=[c_x\, c_y\, c_z]^T\) is the center of the detected object in \(\mathcal{F}_C\), and \(c_u,c_v,f,\alpha\) are the pre-known intrinsic camera parameters. The observation \(\bm{p}\) can be transformed to the \(\mathcal{F}_S\) frame by utilizing knowledge of the object plane direction, and the heading of the UAV. Such a transformation allows us to incorporate depth measurements, reported in \(\mathcal{F}_S\). This can be achieved by using:
\begin{equation}
\label{eq:transform_camera_to_control}
    {}^S\bm{c}={}^Sd({}^S_B{T}\,{}^B_C{T}\,{}^C\bm{p})
\end{equation}
where \({}^Sd\) is a depth measurement projected on \(\bm{s}_x\) axis. Once Eq. \eqref{eq:transform_camera_to_control} is evaluated and \(\bm{r}_c\) is chosen, the error vector in Eq. \eqref{eq:error_function} is known and we can design a controller to minimize it.

\subsection{Sensors Setup and Calibration}
For both the normal camera (we used Intel D435i updating at 60Hz) and the event camera (we used iniVation DAVIS346 updating at 100Hz) we performed intrinsic camera parameters calibration using a checker board. The inertial measurement unit (IMU) we used is Xsens MTi-670 updating at 200 Hz where we also calibrate for gyro drift prior to every flight. We assume a simplified accelerometer error model, and we use position measurements to calibrate accelerometer bias during hovering, assuming uncorrelated 3-axis measurements. The static transformation \({}^B_C{T}\) was found using a CAD model.
We found temporal calibration between the IMU and the camera measurements is the most important. Without temporal synchronization, errors as large as 0.5 m were observed when servoing at around 4 m distance from the object, whereas after synchronization this went down to around 0.05 m. ROS package \emph{messagefilter} was used for the temporal synchronization of sensor measurements.

\subsection{Design of Kalman Filter}
\begin{figure*}
    \centering
    \includegraphics[width=0.9\textwidth]{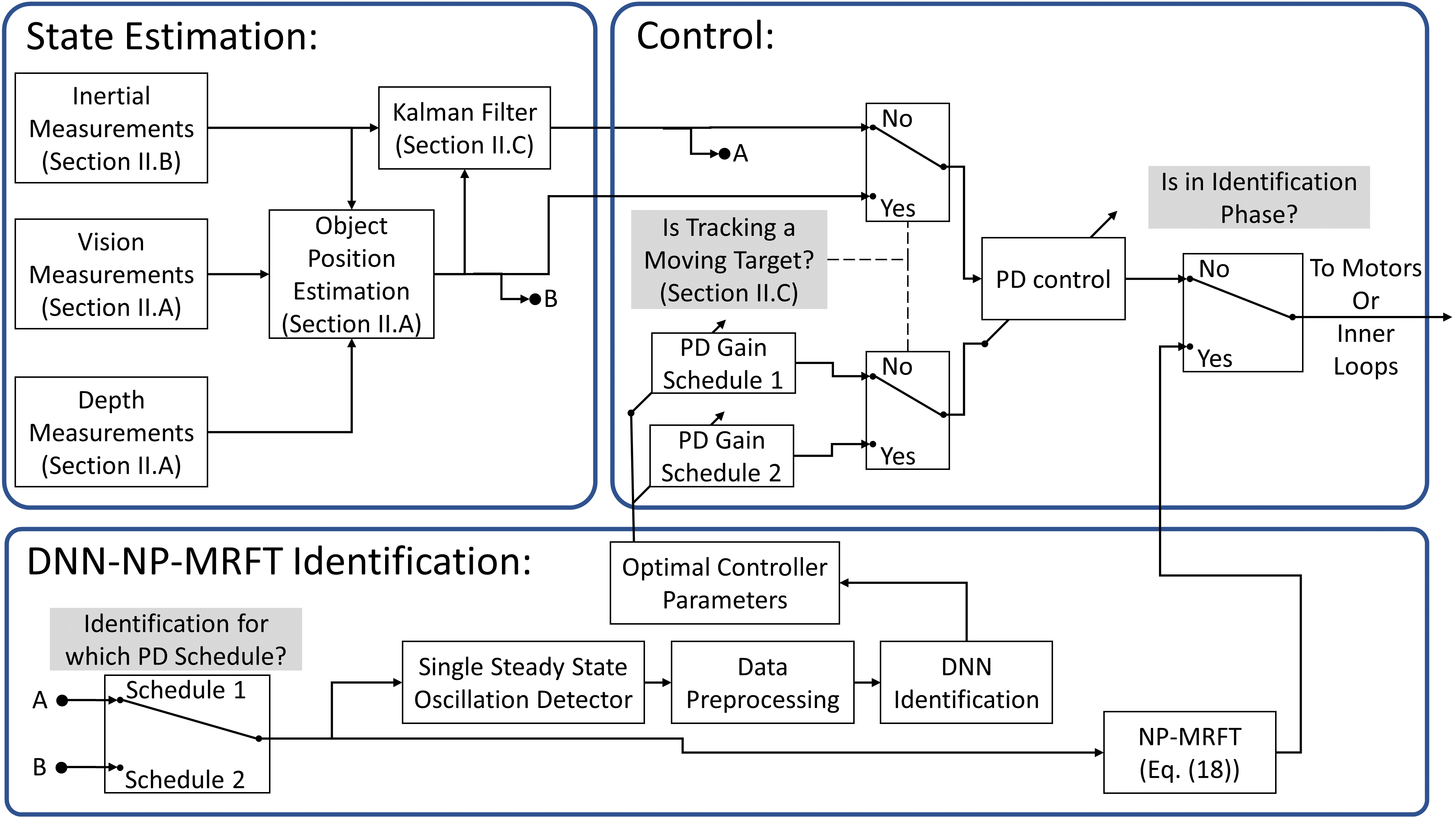}
    \caption{Overall identification structure for visual servoing. Controller Schedule 1 is used for the case when a KF is used to fuse IMU measurements and vision measurements. Controller Schedule 2 is used when vision based measurements are used without being fused with the IMU measurements.}
    \label{fig:identification_structure}
\end{figure*}

We used a KF to obtain smoother and higher rate \({}^S\bm{e}\) and \({}^S\bm{\dot{e}}\) estimates. A decoupled kinematic model was used to design the filter, which separates the rotational and linear motion estimates. The decoupling allows the design of three independent KF's, one along each inertial axis. The filtered states for a single KF are the position \(p\), velocity \(v\), and an estimate of the acceleration bias \(a_b\) which was added to adjust for gravity. The prediction model used is:
\begin{multline}
   \bm{x}_{k+1} =  
   \begin{bmatrix}
   p_{k+1}  \\ v_{k+1}  \\ a_{b, k+1} 
   \end{bmatrix}= 
   \begin{bmatrix}
   1 & \Delta t & -\Delta t^2 \\
   0 & 1 & -\Delta t \\
   0 & 0 & 1 
   \end{bmatrix} \bm{x}_k + 
   \begin{bmatrix}
   \Delta t^2 \\
   \Delta t \\
   0
   \end{bmatrix} u_k\\
   +\mathcal{N}(0, \sigma_p)
\end{multline} 
where \(\bm{x}\) is the state vector, the subscript \(k\) is an integer multiple of the step time \(\Delta t\), \(u\) is the control signal, and \(\mathcal{N}(0, \sigma_p)\) is a zero mean normally distributed additive noise with variance \(\sigma_p\). The control signal is obtained from rotating the accelerometer measurement into the inertial frame. The aforementioned model assumes an exact knowledge of the rotation of the UAV, and an identical additive noise profile of the three accelerometer axes, such that \(u\) is not scaled improperly due to misalignment, and \(\sigma_p\) does not change as the UAV rotates. The correction step uses the position estimate form the camera, described in section \ref{sec:obj_det}, with the addition of measurement noise \(\mathcal{N}(0, \sigma_c)\).
The KF cannot be used when the tracked object starts moving as such movements are not captured by the inertial sensors, thus introducing inconsistency in the filter states. Consequently, the KF can be only used when the object is stationary, and when it moves we have to solely depend on camera measurements. This is reflected in the state estimation part of Fig. \ref{fig:identification_structure} where a large error in the Kalman Filter prediction step triggers a change in the controller schedule. The state estimator dynamics will be picked up in the identification phase, and hence more sophisticated nonlinear estimators like those suggested in \cite{ALSHARMAN2018,ALSHARMAN2020} can still be used in this approach. Furthermore, a recent approach which uses the DNN-MRFT framework to design a dynamic based KF can be used to provide state estimates at higher rates \cite{wahbah2021dynamic}.

\section{Identification and Controller Tuning}
This work proposes DNN-NP-MRFT for visual servoing of multirotor UAVs, as noisy sensor measurements cannot be handled directly with the existing DNN-MRFT formulation suggested in \cite{ayyad2020real}. The noise mainly affects the switching behaviour of MRFT. The use of filters in the identification phase introduces lag to the system dynamics, which would negatively affect the closed loop performance at the control phase. Some nonlinear techniques for the avoidance of false MRFT switching like inhibited switching for certain period of time, or the use of two independent relays for the positive and negative switching phases are described in \cite{Boiko2013book}. But such false switching prevention techniques are not suitable when the MRFT's \(\beta\) parameter is negative with a large magnitude. UAV visual servoing systems usually suffer from non-Gaussian noise, usually caused by pixelation errors and coupling between IMU and vision measurements, while still requiring large negative magnitude of the MRFT parameter \(\beta\). For example, Fig. \ref{fig:event_camera_swing} shows the response of UAV altitude dynamics when relative altitude given by the event camera (without IMU fusion), for DNN-NP-MRFT. Fig. \ref{fig:event_camera_swing} shows how DNN-NP-MRFT prevents false switching at multiple instances where the methods described in \cite{Boiko2013book} would fail.

Before discussing the details for robust MRFT formulation, we first introduce the dynamic model that is used for DNN-NP-MRFT identification and controller synthesis.

\subsection{Linearized Dynamic Models}
A multirotor UAV is a rigid body in \(\mathbb{R}^3\), having 6 degrees-of-freedom (DOF) and subject to forces and torques in \(\mathbb{R}^3\). It is assumed that the multirotor UAV Center of Gravity (CoG) is coincident with the IMU center, intersects the propeller's plane, and is also coincident with the body frame \({}^B\mathcal{F}\) origin.We also we assume the UAV body to be symmetric around all axes. Forces and torques are applied to the rigid body by the actuators, where they are assumed to be mapped from the electronic speed controllers (ESCs) inputs by the following linear propulsion system dynamics:
\begin{equation}\label{eqn:actdyn}
    G_{prop}(s) = \frac{K_{eq}e^{-\tau s}}{T_{prop}s+1}
\end{equation}
where \(K_{eq}\), \(\tau\), and \(T_{prop}\) are lumped parameters describing overall system dynamics \cite{Cheron2010}. Following the modeling approach detailed in \cite{Chehadeh2019,ayyad2021tcst} we linearize the nonlinear dynamics of multirotors around the hover state, and cascade the actuator dynamics to obtain the following attitude and altitude dynamics:
\begin{equation}\label{eqn:innerloopdyn}
    G_{inner}(s) = \frac{K_{eq}e^{-\tau_{in} s}}{s(T_{prop}s+1)(T_1s+1)}
\end{equation}
and the lateral motion dynamics are given by:
\begin{equation}\label{eqn:outerloopdyn}
    G_{lat} (s) = \frac{K_{eq}e^{-\tau_{out} s}}{s^2(T_{prop}s+1)(T_1s+1)(T_2s+1)},
\end{equation}
The overall lumped dynamics presented in Eqs. \eqref{eqn:innerloopdyn} and \eqref{eqn:outerloopdyn} account for sensor dynamics by including time delay as a parameter. Sensor dynamics are mainly exhibited by a time delay, and therefore would be picked up by the identification algorithm as illustrated in \cite{ayyad2020real,ayyad2021tcst}.

\subsection{Mitigation of noise effects in the MRFT tuning}
\label{sec:robust_mrft}

The input to the DNN is a periodic signal generated as a self-oscillation by the MRFT, which is a discontinuous control algorithm given by the following equations:
\begin{multline}
\label{eq_mrft_algorithm}
u_M(t)=\\
\left\{
\begin{array}[r]{l l}
h\;\; \text{if} \; e(t) \geq b_1\; \lor\; (e(t) > -b_2 \;\land\; u_M(t-) = \;\;\, h)\\
-h\;\; \text{if}\; e(t) \leq -b_2 \;\lor\; (e(t) < b_1 \;\land\; u_M(t-) = -h)
\end{array}
\right.
\end{multline}
where \(b_1=-\beta e_{min}\) and \(b_2=\beta e_{max}\), \(e_{max}>0\) and \(e_{min}<0\) are the last maximum and minimum values of the error signal after crossing the zero level  respectively, and \(u_M(t-)=lim_{\epsilon\rightarrow0^+ }u_M(t-\epsilon)\) is the last control output. Initially, we set \(e_{max}=e_{min}=0\). \(\beta\) is a tunable parameter of the MRFT that defines the phase of the excited oscillations and is bounded by \(-1<\beta<1\). Using the describing function (DF) method, we can show that the self-excited oscillations satisfy the harmonic balance equation \cite{atherton1975}:
\begin{equation}
\label{eq_hb}
N_d(a_0)G(j\Omega_0)=-1
\end{equation}
The DF of the MRFT is presented in \cite{Boiko2013book} as follows:
\begin{equation}\label{eq_mrft_df}
N_d(a_0)=\frac{4h}{\pi a_0}(\sqrt{1-\beta^{2}}-j\beta)
\end{equation}
The optimal value of \(\beta\) for the altitude loop \(\beta_{z0}=-0.72\) is found in \cite{ayyad2021tcst}, and for lateral control loop the optimal value \(\beta_{x0}\) changes based on the identified attitude loop dynamics. This follows the general methodology of finding the optimal value of the MRFT parameter for the quadrotor tests presented in \cite{Boiko2013book}. It is also shown in \cite{ayyad2021tcst} that the periodic motions produced in the MRFT, for the multirotor UAV dynamics are orbitally stable. Orbital stability is still guaranteed for the case of visual servoing, because the effect of the visual servoing is in the higher delay introduced in the loop; the system structure remains the same.

\begin{figure*}
    \hspace*{-1.1cm}  
    \centering
    \includegraphics[width=0.9\textwidth]{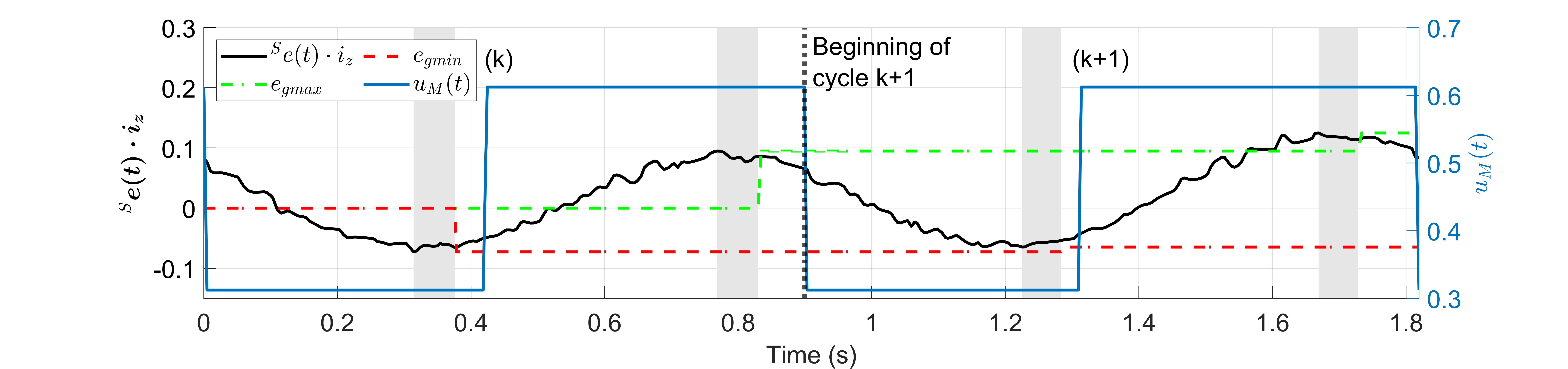}
    \caption[width=0.8\textwidth]{NP-MRFT induced oscillations on the altitude loop for the case of event camera based servoing without KF. The shaded gray area is due to \(\tau_{obs}\) switch inhibition parameter and is defined after \(t_{gmax}^{(k)}\) and \(t_{gmin}^{(k)}\) occurrence. The local peaks and anti-peaks introduced to the main periodic signal by noise would otherwise cause false MRFT switching. The shaded gray area does neither overlap the transitions between \(-h\) to \(h\) nor \(h\) to \(-h\), which indicates that NP-MRFT does not alter the true switching frequency.}
    \label{fig:event_camera_swing}
\end{figure*}

The use of a DNN allows one to rectify the inaccuracies existing due to the use of the assumption concerning the sinusoidal shape of the MRFT oscillation. The DNN can provide an accurate map for any shape of the periodic MRFT oscillation (error signal). In the systems given by the models used for the multirotor dynamics description, the shape of the error signal largely depends on the relative degree and the time delays present in the underlying model \cite{rehan2021optimal}. Indeed, the DF of the MRFT algorithm given by  \eqref{eq_mrft_df} assumes a sinusoidal error signal, whereas the switching given by \eqref{eq_mrft_algorithm} is based on the detected extrema in the error signal \(e_{max}\) or \(e_{min}\) under the sinusoidal shape assumption. Perturbations of the extrema values due to the use of noisy vision sensors would cause errors in switching phases. Orbital stability of the oscillations under the MRFT for the considered model structure was demonstrated in \cite{ayyad2021tcst}, which suggests that the effect of such perturbations would eventually vanish. However, the presence of noise may affect the MRFT based identification and tuning in many ways. Four different effects can be considered as altering the test results. These are:
\begin{enumerate}
\item{Advance switching due to the presence of noise;}
\item{Alteration (increase) of the peak values $e_{max}$ and $|e_{min}|$ due to the addition of noise;}
\item{Local peak detection and computation of the next switching condition based on the incorrectly identified peak value;}
\item{Multiple switching of the relay function in Eq. \eqref{eq_mrft_algorithm} forth and back (chattering).}
\end{enumerate}
We address the effects listed under items 1 and 2 using the following model of noise:
\begin{equation}
    \label{eq_noise_model}
    e(t)=a_0\sin{(\Omega_0 t)}+a_n\sin{(\Omega_n t+\psi_n)}
\end{equation}
where $a_0$ is the true amplitude of the MRFT oscillations unaffected by noise, 
$\Omega_0$ is the frequency of the MRFT oscillations, $a_n$ , $\Omega_n$ and $\psi_n$ are the amplitude, frequency and phase of noise, respectively. The amplitude of noise is assumed to satisfy: $a_n << a_0$ (for the practical consideration, the amplitude of noise can be up to 20\% of $a_0$), and the frequency of noise is $\Omega_n >> \Omega_0$. It is noted that $\Omega_n$ and $a_n$ does not have to be constants and may be fluctuating. Therefore, the proposed model of noise assumes random fluctuations of the amplitude and frequency and uses certain averages values.

According to \eqref{eq_mrft_algorithm}, the switching conditions are established on every half-period of the MRFT oscillation by computing the values of $b_1=-\beta e_{min}$ for the switching from $-h$ to $+h$, and $b_2=\beta e_{max}$ for the switching from $+h$ to $-h$. Switching occurs when the error signal $e(t)$ becomes equal to $b_1$ or $b_2$ (along with some additional conditions given in \eqref{eq_mrft_algorithm}). With satisfaction of the condition $\Omega _0<<\Omega _n$, at the error signal affected by noise, each switching would occur when
\begin{equation}
\nonumber
e(t)=a_0  \sin(\Omega_0 t)+a_n=b_1
\end{equation}
for the switching from $-h$ to $+h$, and 
\begin{equation}
\nonumber
e(t)=a_0  \sin(\Omega_0 t)-a_n=-b_2
\end{equation}
for the switching from $+h$ to $-h$. The presence of $a_n$ in these equations causes an advance switching of the function in \eqref{eq_mrft_algorithm}, so that the effective hysteresis values for the switching function \eqref{eq_mrft_algorithm} are $b_1^*=b_1-a_n$ and $b_2^*=b_2-a_n$. Therefore, the switching occurs when the unaffected by the noise error signal becomes equal to $b_1^*$ and $-b_2^*$.

The peak value is also affected by the noise, so that the measured peak values are $e_{max}=a_0+a_n$ and $e_{min}=-(a_0+a_n)$. To account for the effect of noise, the following two measures are proposed: a correction to the switching condition and a correction to the peak values. With symmetric $e(t)$, the values of $b_{1,2}$ used in \eqref{eq_mrft_algorithm} are adjusted as $b_{1,2}=b_{nom1,2}+a_n$, where $b_{nom1,2}$ are the nominal values computed as $b_{nom1,2}=\beta a_0$. On the other hand, the amplitude of the oscillation not affected by noise is $a_0=e_{max}-a_n$. Putting these relationships together yields:
\begin{equation}
\label{eq_modif_sw2}
b_2=\beta (e_{max}-a_n)+a_n=\beta e_{max}+a_n(1-\beta)
\end{equation}
\begin{equation}
\label{eq_modif_sw1}
b_1=-\beta (e_{min}+a_n)+a_n=-\beta e_{min}+a_n(1-\beta)
\end{equation}
The first term in \eqref{eq_modif_sw2}, \eqref{eq_modif_sw1} provides the switching value for the MRFT algorithm \eqref{eq_mrft_algorithm} without account of noise, and the second one is the correction term accounting for noise. The noise signal is produced through the high-pass filtering of the overall signal, with its amplitude measured as a certain average amplitude.


The above-given noise associated corrections are valid only if the noise  neither alters the switching phase nor the frequency of the MRFT induced oscillations, which was an assumption. To prevent the effects described in items 3 and 4 above from altering the frequency, some additional measures must be taken. Namely, the algorithm must be able to distinguish in real time between a local peak caused by the noise component and the peak of the MRFT oscillation. This is particularly important for the large negative values of $\beta$, as according to \eqref{eq_mrft_algorithm}, after reaching the maximum $e_{max}$, the switch is supposed to occur when the error becomes equal to $-b_2 =-\beta e_{max}$, which may be quite close to $e_{max}$. To distinguish between a local and a global maximum in real time, it is necessary to observe the evolution of $e(t)$ for some time after reaching a maximum to see whether it continues to decrease or increases to a new maximum value. If the frequency of noise $\Omega_n$ (or the minimum frequency of noise) is known, then this observation time can be selected as $\tau_{obs}=T_n=\frac{2 \pi}{\Omega_n}$ or higher than $T_n$.
Let us introduce the real-time current maximum of the error on $k$-th period of the MRFT oscillation as follows:
\begin{equation}
\label{eq_cur_max}
e^{(k)}_{max}(t)=
\left\{
\begin{array}[r]{l l}
\max_{t \in [t^{(k)}_{0+},t]} e(t) \; \text{if} \; e(t)>0\\
0 \; \text{otherwise}
\end{array}
\right.
\end{equation}
where 
$t^{(k)}_{0+}:=\{ t \;|\; e(t^{(k)}_{0+})=0 \; and \; e(t)>0 \; \forall \; t>t^{(k)}_{0+}\}$ is the time of the last zero crossing by increasing $e(t)$. Similarly, we define the the real-time current minimum of the error on $k$-th period of the MRFT oscillation:
\begin{equation}
\label{eq_cur_min}
e^{(k)}_{min}(t)=
\left\{
\begin{array}[r]{l l}
\min_{t \in [t^{(k)}_{0-},t]} e(t) \; \text{if} \; e(t)<0\\
0 \; \text{otherwise}
\end{array}
\right.
\end{equation}
where 
$t^{(k)}_{0-}:=\{ t \;|\; e(t^{(k)}_{0-})=0 \; and \; e(t)<0 \; \forall \; t>t^{(k)}_{0-}\}$ is the time of the last zero crossing by decreasing $e(t)$.
With the current maximum and minimum defined, we can define the global maximum and global minimum on $k$-th period of an MRFT oscillation as follows:
\begin{equation}
\label{eq_glob_max}
e^{(k)}_{gmax}(t)=
\left\{
\begin{array}[r]{l l}
e^{(k)}_{max}(t) \; \text{if} \; e^{(k)}_{max}(t)=e^{(k)}_{max}(t-\tau_{obs})\\
e^{(k-1)}_{gmax}(t) \; \text{otherwise}
\end{array}
\right.
\end{equation}
and
\begin{equation}
\label{eq_glob_min}
e^{(k)}_{gmin}(t)=
\left\{
\begin{array}[r]{l l}
e^{(k)}_{min}(t) \; \text{if} \; e^{(k)}_{min}(t)=e^{(k)}_{min}(t-\tau_{obs})\\
e^{(k-1)}_{gmin}(t) \; \text{otherwise}
\end{array}
\right.
\end{equation}

Therefore, the algorithm accepts the highest value as a global maximum if the maximum value does not change upon adding an observation interval to the interval within which the maximum value is searched. The same algorithm applies to finding the global minimum.

Additionally, the times corresponding to the global maximum and the global minimum on $k$-period are determined within the same algorithm:
$t^{(k)}_{gmax}:=\{ t \;|\; e(t)=\max_{t \in [t^{(k)}_{0+},t^{(k)}_{gmax}+\tau_{obs}]} e(t)$
and
$t^{(k)}_{gmin}:=\{ t \;|\; e(t)=\min_{t \in [t^{(k)}_{0-},t^{(k)}_{gmin}+\tau_{obs}]} e(t)$. The values of \(t^{(k)}_{gmax}\) and \(t^{(k)}_{gmin}\) are initialized properly so that the periods \(t \in [t^{(k)}_{0+},t^{(k)}_{gmax}+\tau_{obs}]\) and \(t \in [t^{(k)}_{0-},t^{(k)}_{gmin}+\tau_{obs}]\) are properly defined. The possibility of chattering, which is listed above as item 4, is prevented from happening through inhibiting of switching from the moment of the previous switch of the algorithm from $-h$ to $h$ (or $h$ to $-h$) to the time $t^{(k)}_{gmax}+\tau_{obs}$ (or $t^{(k)}_{gmin}+\tau_{obs}$), respectively.

Thus we augment the MRFT Eq. \eqref{eq_mrft_algorithm} with the conditions and equations aimed at addressing the noise, as follows:
\begin{strip}
\begin{equation}
\label{eq_np_mrft_algorithm}
u^{(k)}_M(t)=\left\{
\begin{array}[r]{l l}
h\;\;\;\; \text{if} \; (e(t) \geq b^{(k)}_1 \land t>t^{(k)}_{gmin}+\tau_{obs}) \; \lor\; (e(t) > -b^{(k)}_2 \;\land\; u^{(k)}_M(t-) = \;\;\, h )\\
-h\;\; \text{if}\; (e(t) \leq -b^{(k)}_2 \land t>t^{(k)}_{gmax}+\tau_{obs}) \;\lor\; (e(t) < b^{(k)}_1 \;\land\; u^{(k)}_M(t-) = -h )
\end{array}
\right.
\end{equation}
\end{strip}
where $b_1$ and $b_2$ are now defined by the equations:
\begin{equation}
\label{eq_modif_sw2a}
b^{(k)}_2=\beta e^{(k)}_{gmax}+a_n(1-\beta)
\end{equation}
\begin{equation}
\label{eq_modif_sw1a}
b^{(k)}_1=-\beta e^{(k)}_{gmin}+a_n(1-\beta)
\end{equation}
and \eqref{eq_glob_max} and \eqref{eq_glob_min}.
%

We shall further refer to the algorithm \eqref{eq_np_mrft_algorithm} as the Noise-protected MRFT (NP-MRFT). With the NP-MRFT algorithm it was possible to avoid the undesired effect described in item (3) above.
%
However, there is a cost that is paid for the noise protection feature offered by the NP-MRFT, due to the necessity of having an observation time $\tau_{obs}$ after detecting a maximum/minimum, which is a limiting factor on the range of possible $\beta$. Based on the assumption of the sinusoidal shape of the NP-MRFT self-excited oscillations, the minimum realizable switching phase \(\beta_{min}\), is given by:
\begin{equation}
    \beta_{min}=-1+\sin\frac{\tau_{obs}}{T_0}
\end{equation}
where \(T_0=\frac{2\pi}{\Omega_0}\) is the period of the NP-MRFT oscillations unaffected by noise as shown in Eq. \eqref{eq_noise_model}. For the case \(\beta\leq\beta_{min}\) the DF of the NP-MRFT is the same as of MRFT, but would be otherwise not applicable and not suitable for DNN-NP-MRFT based identification.

The values of \(T_n\) and \(T_0\) are obtained experimentally, are specific for every sensor used, and to be found only once. To obtain \(T_n\) in-flight, we use a high-pass filter with a suitable cut-off frequency combined with a period estimation through consecutive peaks detection. The value of \(T_0\) is estimated from the measurement of the NP-MRFT induced steady-state periods. We found the value of \(\beta_{min}\) to vary significantly based on the sensor's configuration in use. For example, the normal camera with KF has a ratio of \(\frac{T_n}{T_0}=0.015\) which results in \(\beta_{min}=-0.906\)). \(\beta_{min}\) for most cases was smaller than \(\beta^*_{z0}\) or \(\beta^*_{x0}\) which allowed us to realize the optimal value of the phase of the test. Fig. \ref{fig:event_camera_swing} shows how NP-MRFT is preventing false switching from occurring without altering the switching phase. In the case of thermal camera when KF was not used, realizing stable oscillations at \(\beta^*_{z0}\) was not possible due to the large value of \(\beta_{min}\), and hence the value of \(\beta_{z0}\) was shifted from the optimal test phase \(\beta^*_{z0}\).

\subsection{Identification with DNN-NP-MRFT}
The DNN part used for the identification of unknown dynamics remains the same as DNN-MRFT presented in \cite{ayyad2020real,ayyad2021tcst}. For the convenience of the reader, we provide in the following a brief discussion on data generation, training, and stability aspects of DNN-NP-MRFT.

Using a classification approach for identification provides several advantages like the computational efficiency in training, and the ability to provide optimal controller parameters in a lookup table for real-time tuning. Controller parameters are tuned to minimize the ISE cost for a step reference. Theoretically, it is possible with DNN-NP-MRFT to obtain, for the same system, two different sets of the lumped system parameters that would result in the same tuning of controller parameters, and hence the same closed loop system performance. This is because DNN-NP-MRFT benchmarks identification accuracy based on the closed loop system performance. In the DNN-NP-MRFT based approach, the relative sensitivity function is used as a distance measure in the system parameters' space  \cite{rohrer1965sensitivity}. The relative sensitivity function is defined by:
\begin{equation}\label{eq_performance_deterioration}
J_{ij} = \frac{Q(C_i, G_j) - Q(C_j, G_j)}{Q(C_j, G_j)} \times 100 \%
\end{equation}
where \(J_{ij}\) represents the degradation in performance due to applying controller \(C_i\), which is the optimal controller for the process \(G_i\) and a sub-optimal controller for the process \(G_j\). \(Q\) is a function that calculates the ISE of the step response of the closed loop system. Note that \(J_{ij}\neq J_{ji}\) so we define \(J_{(ij)}=max(J_{ij},J_{ji})\). 

The value of \(J_{(ij)}\) is used to properly discretize the system parameter's space within the defined possible range of parameters. Due to the gain scale property, the unknown system gain can span the range from zero to infinity. The unknown time parameters are chosen based on the modeling of UAV parameters in \cite{Chehadeh2019}, and these ranges were found to be \(T_{prop}\in[0.015,0.3]\), \(T_{1}\in[0.2,2]\), \(T_{2}\in[0.2,6]\), \(\tau_{in}\in[0.0005,0.03]\), and \(\tau_{out}\in[0.0005,0.15]\). For the altitude case, the loop delay \(\tau_{in,alt}\in[0.0005,0.06]\) was increased due to the dependency of altitude estimation on slow vision dynamics. For each of the discretized systems within the specified range of parameters, MRFT was used to generate training data based on numerical simulations. To avoid overfitting, the training data is augmented with white Gaussian noise and random bias that is introduced to the simulated dynamics.

The set of DNNs we used are as follows. First, a DNN for attitude loops with 48 output classes (where each output corresponds to a discretized process). Another DNN was used for altitude identification with 208 output classes. There are more outputs for the altitude case as \(\tau_{in,alt}\), the time delay range, was widened compared with the attitude case. As the lateral loop dynamics depend on the underlying attitude dynamics, we used a set of 48 DNNs each corresponding to a discretized attitude class. Each lateral dynamics DNN has an average of 15 output classes.

The cost function used for DNN-NP-MRFT training takes into account the value of \(J_{(ij)}\) that resulted from misclassification. The DNN modified softmax function with cross-entropy loss introduced in \cite{ayyad2020real}:
\begin{equation} \label{eq_modified_softmax}
p_{i} = \frac{e^{J_{il}\cdot a_{i}}}{ \sum_{j=1}^{N}  e^{J_{jl}\cdot a_{j}}}
\end{equation} 
where \(l\) represents the label index available in the training and:
\begin{equation} \label{eq_modified_softmax_backprop}
\frac{\partial L}{\partial a_{i}} = J_{il} \times (p_{i} - y_{i})
\end{equation}
is the back-propagation term for the one-hot encoded vector. The involvement of the \(J_{il}\) in training the network caused the DNN to avoid misclassifications that would result in significant performance drop or instability as illustrated by the Monte Carlo analysis in \cite{ayyad2020real}. 

The selected DNN structure consists of two hidden layers of size 3000 and 1000 neurons respectively, with Rectified Linear Units (ReLU) activation functions. ADAM optimizer was used during the training. Each DNN training took one to two minutes. The DNN prediction time is just a few milliseconds on modern embedded processors (such as Broadcom BCM2837 which is used on the Raspberry Pi 4), thus enabling real-time identification and tuning.

\renewcommand{\arraystretch}{2}
\begin{table}[t]
\setlength\tabcolsep{4 pt} 
\caption{Identified altitude parameters and controller gains resulted from DNN-NP-MRFT. Also, step ISE cost are provided for simulation and experimentation based on the identified parameters. The target tracking experiment was not successful with the thermal camera due to the large noise resulting from position differentiation and limited camera FOV.}
    \centering
 \begin{tabular}{|p{0.006\textwidth}|c|>{\centering\arraybackslash}c|>{\centering\arraybackslash}c|>{\centering\arraybackslash}c|>{\centering\arraybackslash}c|>{\centering\arraybackslash}c|>{\centering\arraybackslash}c|>{\centering\arraybackslash}c|} 
 \hline
 \multicolumn{2}{|c|}{Sensor}&  \(T_{prop}\) &  \(T_1\) & \(\tau_{in}\) & \(K_p\) & \(K_d\) & \(Q_{exp}\) & \(Q_{sim}\) \\
 \hline
 \multirow{3}{4em}{\rotatebox[origin=c]{90}{with KF}} & Normal & 0.30 & 0.20& 0.0128 & 1.1766 & 0.3143 & 0.0874 & 0.080 \\
 \cline{2-9} 
 & Event & 0.30 & 0.20& 0.0128 & 1.2993 & 0.3471 & 0.0771 & 0.0698  \\ \cline{2-9} 
 & Thermal & 0.1675 & 0.6523 & 0.0048 & 1.7766 & 0.5352 & 0.0933 & 0.0684 \\
 \hline
 \multirow{3}{4em}{\rotatebox[origin=c]{90}{without KF}} & Normal & 0.1355 & 1.6825 & 0.06 & 0.3739 & 0.1705 & 0.2174 & 0.1629\\
 \cline{2-9} 
 & Event & 0.0135 & 1.6834& 0.0237 & 0.888 & 0.3258 & 0.1293 & 0.1068 \\
 \cline{2-9}
 & Thermal & 0.2766 & 2.0059 & 0.0022 & 0.257 & 0.1435 & - & -\\
 
 \hline
\end{tabular}
\label{tab:ident_results_altitude}
\end{table}

\renewcommand\arraystretch{0.75}
\begin{figure}
  \centering
  \begin{tabular}{@{}c@{}}
    \hspace*{-0.5cm}  
    \includegraphics[width=\columnwidth]{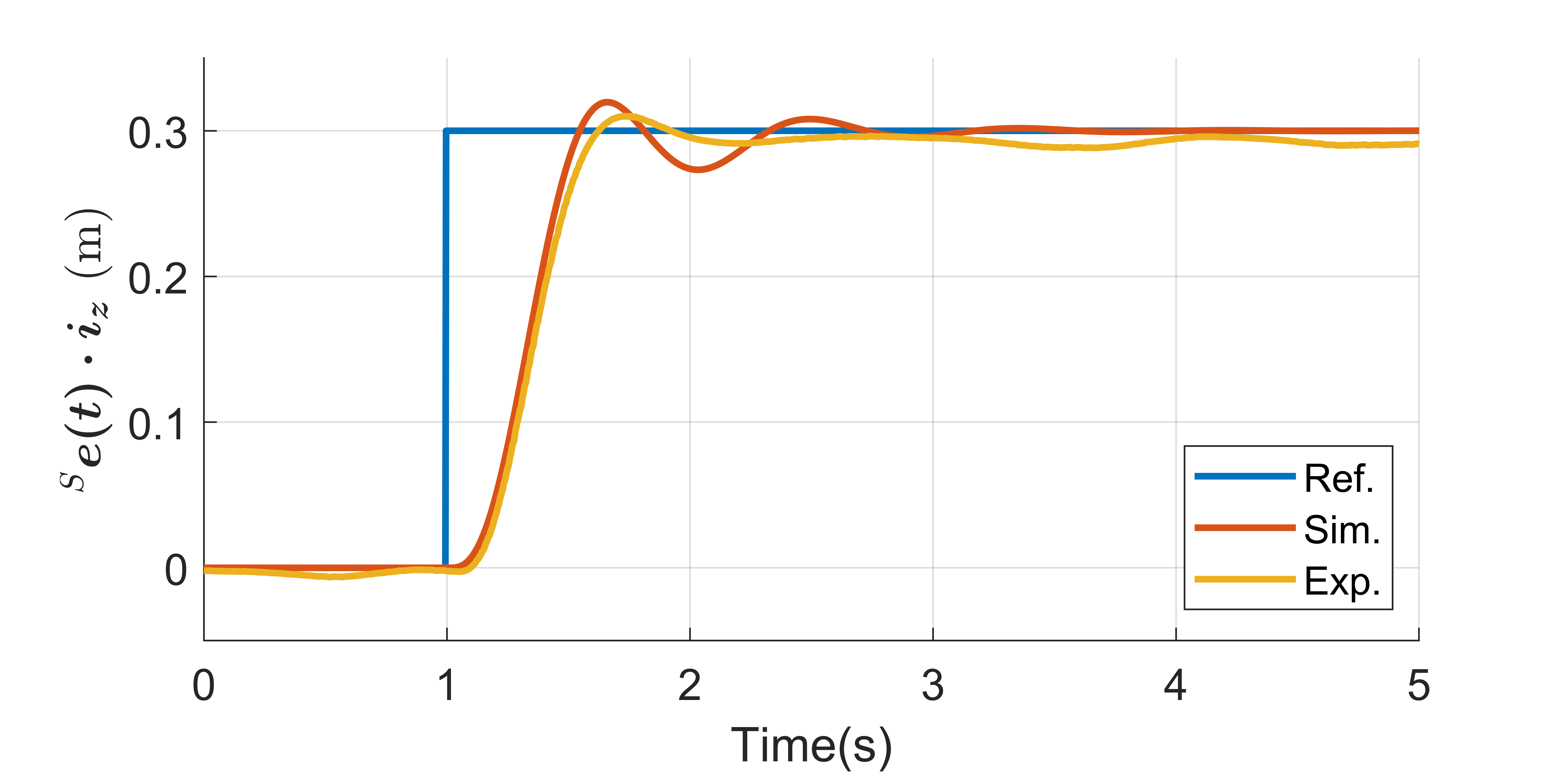}
    \\   \small (a) 
  \end{tabular}
  

  \begin{tabular}{@{}c@{}}
    \hspace*{-0.5cm}  
    \includegraphics[width=\columnwidth]{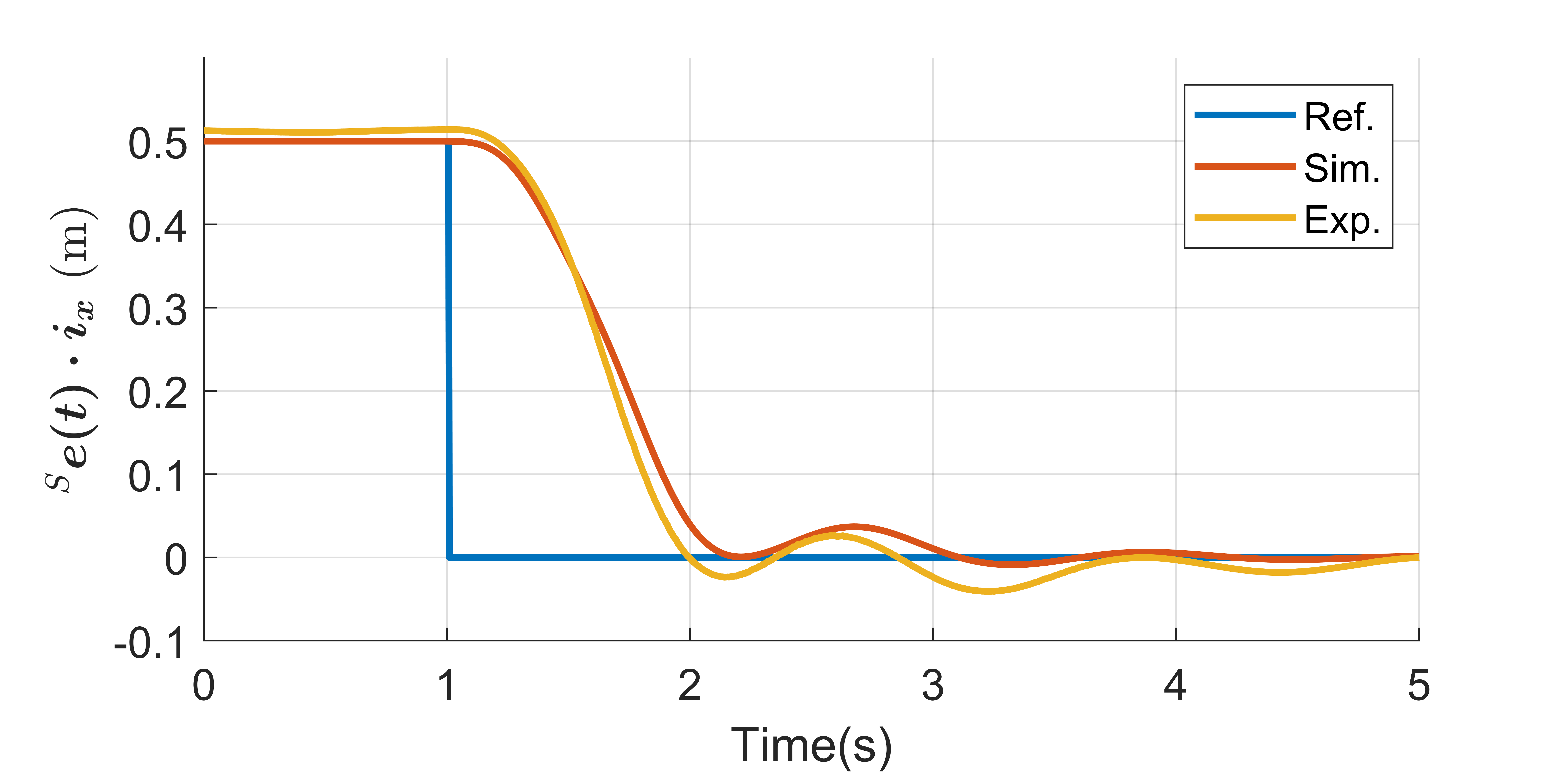}
    \\      \small (b)
  \end{tabular}

  \caption{System response to a step reference change when normal camera was used with a KF to estimate position and velocity. PD parameters used correspond to Schedule 1 in Fig. \ref{fig:identification_structure}.} \label{fig:rgb_step_kf}
\end{figure}

\section{Results}
For all experiments, we have used a hexarotor UAV running on a custom flight controller software. The main loop runs on a Raspberry PI 3B embedded computer with the NAVIO2 extension board. The main control loop runs at 200Hz in a ROS node. Without the camera payload, the hexarotor dimensions are \(111\times100\times27\)cm, its weight is 3.38 kg, and the estimated rotational inertias are \(J_x=0.093,\,J_y=0.089,\,J_z=0.156\) all in \(kg\cdot m^2\). We have used the E600 propulsion system from DJI, which accepts a PWM command with a pulse width in the range of 1ms to 2ms. The saturation values of the propulsion system measured on a bench test are 1165\(\mu\)s and 1878\(\mu\)s. Additionally, a saturation value of \(\pm 0.2617rad\) was applied to the controller output of the lateral motion control loop. Also, we have used a Butterworth second order low-pass filter with a cutoff frequency of 20Hz. All the saturation values and filters used in experimentation are reflected in all our simulations. The video in \cite{oussama_2021_video} shows the experimental tests done as in this paper, and provides a summary of the results.

\renewcommand{\arraystretch}{2}
\begin{table}[t]
\setlength\tabcolsep{4 pt} 
\caption{Identified lateral motion parameters and controller gains from the DNN-NP-MRFT. Also, step ISE cost are provided for simulation and experimentation based on the identified parameters. The target tracking experiment was not successful with the thermal camera due to the large noise resulting from position differentiation and limited camera FOV.}
    \centering
 \begin{tabular}{|p{0.006\textwidth}|c|>{\centering\arraybackslash}c|>{\centering\arraybackslash}c|>{\centering\arraybackslash}c|>{\centering\arraybackslash}c|>{\centering\arraybackslash}c|>{\centering\arraybackslash}c|>{\centering\arraybackslash}c|} 
 \hline
 \multicolumn{2}{|c|}{Sensor}&  \(T_2\) & \(\tau_{out}\) & \(K_p\) & \(K_d\) & \(Q_{exp}\) & \(Q_{sim}\) \\
 \hline
 \multirow{3}{4em}{\rotatebox[origin=c]{90}{with KF}} & Normal & 0.40 & 0.0825& 1.4343 & 0.5662 & 0.2608 & 0.2574 \\
 \cline{2-8} 
 & Event & 3.6789 & 0.01& 1.0155 & 0.5270 & 0.3332 & 0.2556\\ \cline{2-8} 
 & Thermal & 0.40 & 0.30 & 0.9089 & 0.4962 & 0.2523 & 0.3580 \\
 \hline
 \multirow{3}{4em}{\rotatebox[origin=c]{90}{without KF}} & Normal & 3.6789 & 0.01 & 0.9403 & 0.488 & 0.3855 & 0.3449 \\
 \cline{2-8} 
 & Event & 3.6789 & 0.01& 0.8154 & 0.4231 & 0.2479 & 0.1927 \\
 \cline{2-8}
 & Thermal & 0.40 & 0.30 & 0.3585 & 0.1957 & - & -\\
 \hline
\end{tabular}
\label{tab:ident_results_side}
\end{table}

\subsection{Identification Results}
System identification is performed twelve times in total: twice for every control schedule (\ref{fig:identification_structure}) for every sensor used in the two control loops. The results are given in Table \ref{tab:ident_results_altitude} for the altitude control loop, and in Table \ref{tab:ident_results_side} for the lateral motion control loop. Following the homogeneous tuning approach presented in \cite{Boiko2013book}, we have constrained the minimum phase margin to be 20\degree\ for all the tuning cases. We have used PD controllers for all control loops. The PD controller of the lateral motion outer loop provides the set-point for the corresponding inner attitude loop.

The results in Tables \ref{tab:ident_results_altitude} and \ref{tab:ident_results_side} show that when the KF is used the different sensors resulted in similar controller parameters. The closed loop performance for altitude and lateral step tracking is similar for all cameras used despite the differences in the sensor dynamics. This is explained by the fact that the IMU dynamics are dominant in the transient phase. Tables \ref{tab:ident_results_altitude} and \ref{tab:ident_results_side} also show the step costs obtained in the simulations and experiments. It is noted that finding the step cost for the case when the tracked target moves in a step requires some approximations as the target movement is not synced with the on-board measurements. Fig. \ref{fig:rgb_step_kf} shows the step response match between simulations and experiments for the normal camera. The qualitative behaviour (i.e. overshoot, oscillations nature and decay characteristics, etc.) shown in experiments match the simulated results. This is evidence of the correctness of the used model structure. The percentage overshoot (PO) of the simulated system for the normal camera with KF on altitude step is 6.32\% which is comparable with 7.85\% obtained experimentally. The error in rise time \(T_r\) between simulation and experimentation was 13.4\% for the same step case. The error figures were smallest for the normal camera as it has highest resolution and lowest noise.

\renewcommand\arraystretch{0.75}
\begin{figure}
  \centering
  \begin{tabular}{@{}c@{}}
    \includegraphics[width=\columnwidth]{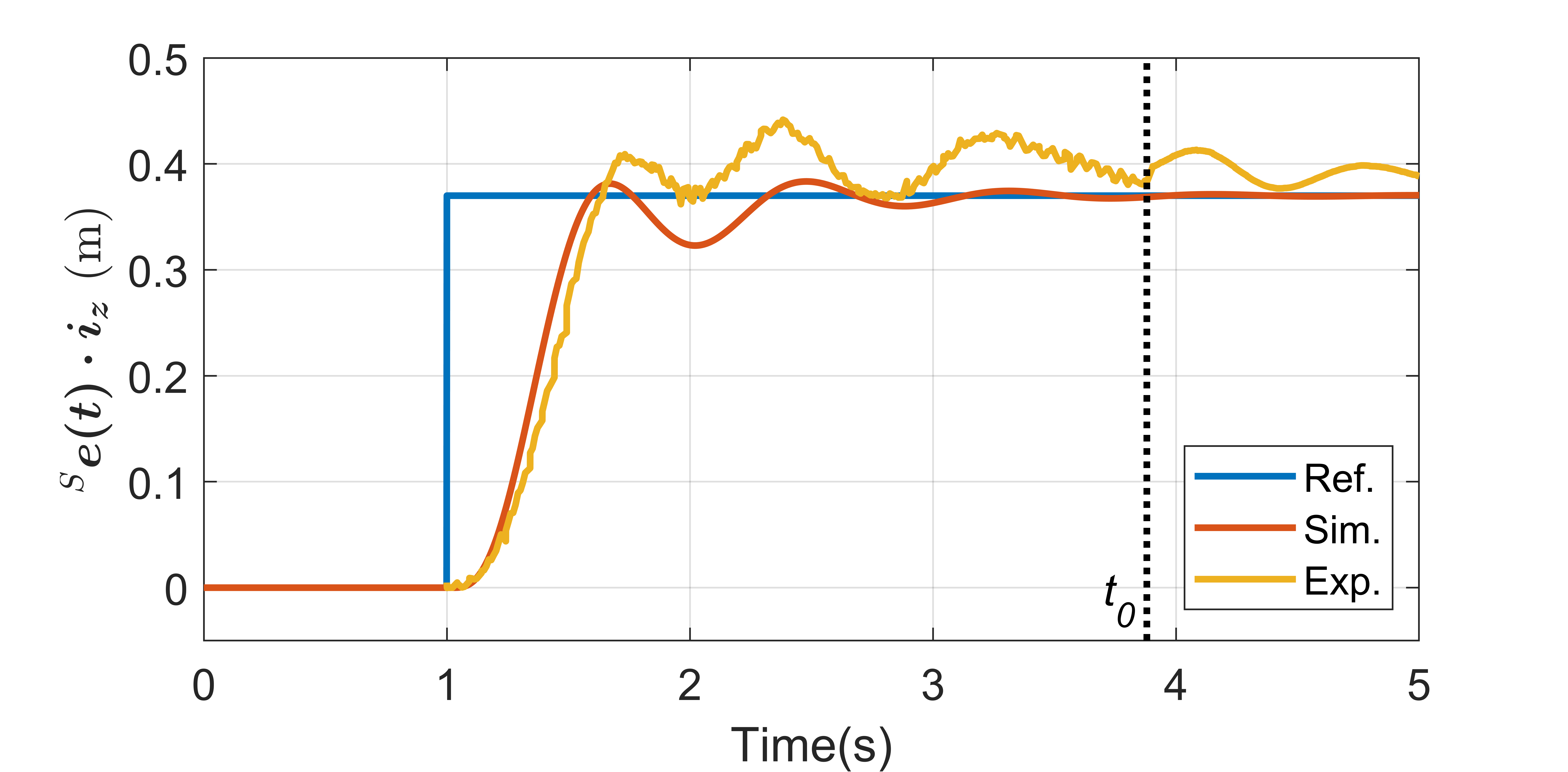}
  \end{tabular}
  

    

  \caption{System response to a step change in object altitude when an event camera was solely used to estimate position and velocity. PD parameters used correspond to Schedule 2 shown in Fig. \ref{fig:identification_structure}. The switch back to the PD Schedule 1 happens at the moment indicated by the vertical dashed line when the system reaches a steady state condition. }\label{fig:det_event_step}
\end{figure}

\subsection{Disturbance Rejection}
We have performed a few disturbance rejection tests. The disturbance types we have considered are tracked target step changes (i.e. vision disturbance), wind disturbances, and pull and release disturbance. Tracking a step change of a target requires the use of control parameters schedule tuned without the KF (i.e. IBVS). Step following is the extreme case of vision target tracking as a step is not Lipschitz and methods similar to those proposed in \cite{thomas2017autonomous,Chaumette2018tracking} for approximating the target motion model would not work under the assumption that the tracked target might move arbitrarily. The identification without the KF shows a significant drop in the system gains for the altitude loops compared to the case when the identification is done with the KF due to higher observed lag in the system as reported in Table \ref{tab:ident_results_altitude}. The system rise time \(T_r\) for the altitude control loop increased from 0.63 to 0.71 for the normal camera, and from 0.49 to 0.64 for the event camera when the KF was not used. Table \ref{tab:ident_results_altitude} shows a better step tracking performance with event camera on the altitude loop. On the other hand, it is interesting to observe a slight drop in the controller performance for the lateral motion control loop case (refer to Table \ref{tab:ident_results_side}) when the KF is not used. This is due to the fact that the faster inner loop dynamics, which use on-board IMU measurements, are dominant in the lateral motion control loop. Fig. \ref{fig:det_event_step} shows the response of a UAV with event camera to a step change in the location of the tracked object.

\renewcommand\arraystretch{0.75}
\begin{figure}
  \centering
  \begin{tabular}{@{}c@{}}
    \includegraphics[width=\columnwidth]{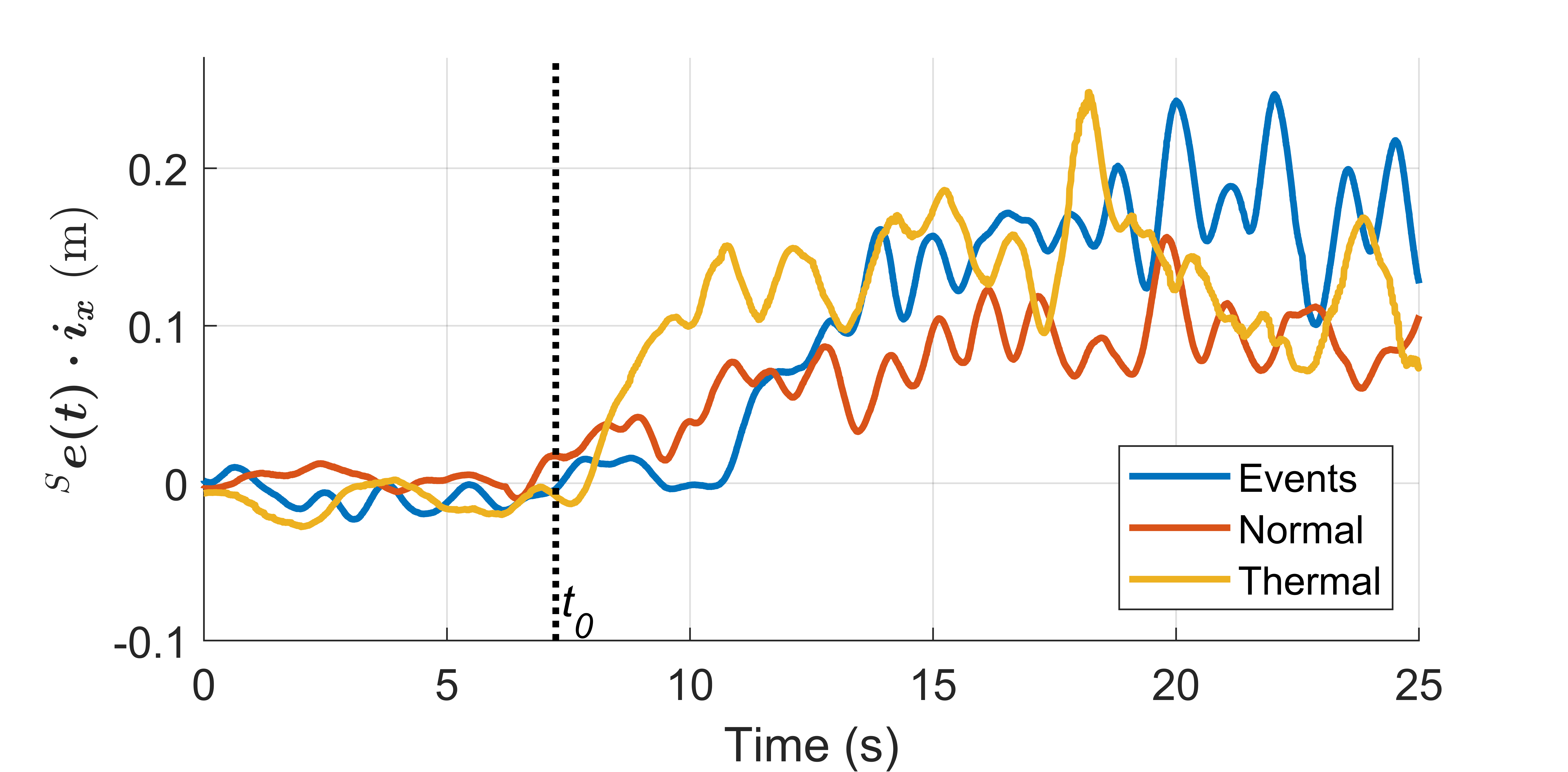}
  \end{tabular}
  

    
  \caption{Response of multirotor for every type of vision sensor used for an external wind disturbance with wind speed of 5 m/s. The fan was turned on at time \(t_0\).}\label{fig:wind_dist_all}
\end{figure}

A steady-state error is expected for the case of wind disturbances when PD controllers are used. Fig. \ref{fig:wind_dist_all} shows the wind disturbance rejection performance where a wind speed of around 5 m/s is applied to the hovering multirotor UAV with the three different sensors. The platform reacts adequately to the disturbance and deviates by less than 18 cm from the hover position for the normal camera case. The platform deviated more from the hover position for the other two sensors due to the lower \(K_p\) values of the controller, which is expected. 

\begin{figure}
  \centering
  \begin{tabular}{@{}c@{}}
    \includegraphics[width=\columnwidth]{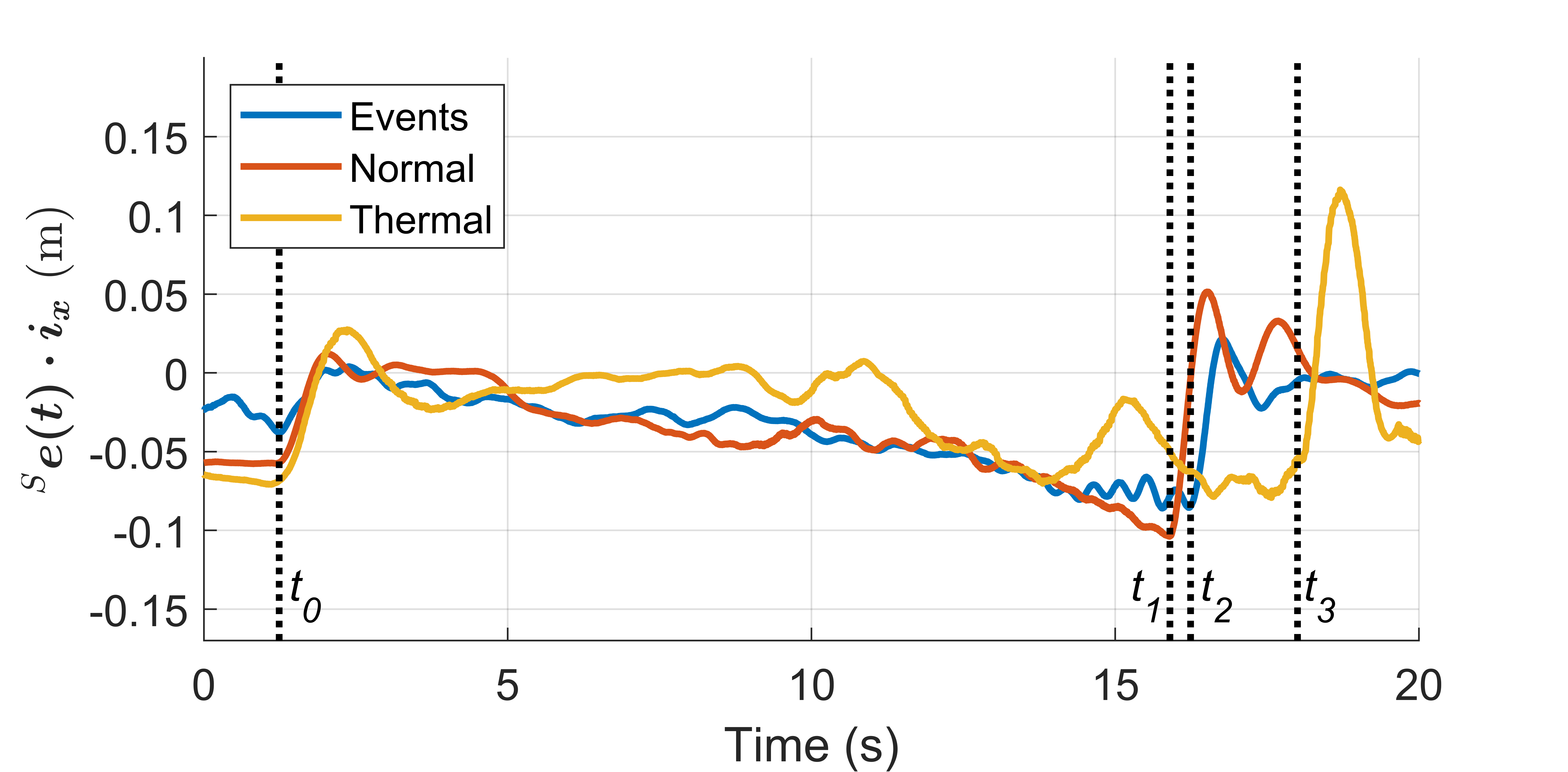}
  \end{tabular}
  

    
  \caption{Response of multirotor for every type of vision sensor used for a pull force of 10 N followed by an instantaneous release. The thread pull starts at time \(t_0\) and the thread cut happened at \(t_1\) for the normal camera, at \(t_2\) for the event camera, and at \(t_3\) for the thermal camera case.}\label{fig:pull_dist_all}
\end{figure}

Pull and release disturbance alters system dynamics in the pull phase as the UAV becomes linked with the human body. At the release instant, the system would have an initial state that needs to be driven to the equilibrium point again. The pull and release disturbance is similar in nature to the case of recovery from collisions or random initial flight condition. We tested the system with all vision sensors by applying a force of 10 N in the pull phase, then releasing the system by cutting the pull thread. Fig. \ref{fig:pull_dist_all} shows the system response to a pull and release disturbance for all the used cameras.

\subsection{Relation to Existing Results}
The low-latency and high throughput of event camera's results in faster dynamics which increases closed loop bandwidth as discussed in \cite{Dimitrova2020}. The match between simulated and experimental results demonstrated in Fig. \ref{fig:rgb_step_kf} for PBVS case, and in Fig. \ref{fig:det_event_step} for IBVS case, confirms the suitability of the third order model of altitude, combined with time delay, to model various visual servoing dynamics. The obtained results in Table \ref{tab:ident_results_altitude} showed a significant drop in optimal performance of altitude dynamics for the case when IMU was not involved in the position estimate, compared with the case where IMU was incorporated in the position estimate. This was not the case for the underactuated lateral loops, as attitude dynamics became dominant in this case. This suggests that the tuning of visual servoing systems, where servoing happens only across lateral dynamics as in the case of landing on a moving ground vehicle \cite{Baca2019landing}, is less sensitive to the visual servoing sensor used.

The percent increase in ISE cost when a normal camera was used, for the experimental case presented in Table \ref{tab:ident_results_altitude}, is 67\% compared with event camera. From a system dynamics perspective, the gains in performance when using event camera's for visual servoing without KF confirms and provides a mean for the quantification of the findings suggested in \cite{Falanga2019fast_too_fast}. This quantification, which is offered by the proposed DNN-NP-MRFT approach, is of importance for the design of UAV platforms requiring high maneuverability, as in the case of sense and avoid \cite{Falanga2019fast_too_fast}, or in prey hunting scenarios \cite{Vrba2022}.

\renewcommand\arraystretch{0.75}

\section{Conclusions}
This paper presents a systematic method to system identification and controller tuning based on DNN-NP-MRFT for visual servoing in the presence of noisy sensor measurements. The proposed approach provide several advantages over existing approaches such as real-time identification, the ability to account for delay dynamics, and systematic tuning. The proposed method was tested on a multitude of sensor and estimator configurations that are common in visual servoing tasks with UAVs. Results are presented to show the match between experimental and simulation results, supporting the conclusion that the proposed approach can be used to predict the system behaviour for different tuning settings and changes in system parameters. The automatically tuned closed loop system demonstrated sufficient and stable performance against different external disturbances. In future work we will investigate the application of the proposed approach to track targets having continuous movements at high speeds, and to further analyse performance mismatch when performing identification with slow update rate sensors.

\bibliographystyle{IEEEtran}
\bibliography{bst/ref_items.bib}

\begin{thebibliography}{10}
\providecommand{\url}[1]{#1}
\csname url@samestyle\endcsname
\providecommand{\newblock}{\relax}
\providecommand{\bibinfo}[2]{#2}
\providecommand{\BIBentrySTDinterwordspacing}{\spaceskip=0pt\relax}
\providecommand{\BIBentryALTinterwordstretchfactor}{4}
\providecommand{\BIBentryALTinterwordspacing}{\spaceskip=\fontdimen2\font plus
\BIBentryALTinterwordstretchfactor\fontdimen3\font minus
  \fontdimen4\font\relax}
\providecommand{\BIBforeignlanguage}[2]{{%
\expandafter\ifx\csname l@#1\endcsname\relax
\typeout{** WARNING: IEEEtran.bst: No hyphenation pattern has been}%
\typeout{** loaded for the language `#1'. Using the pattern for}%
\typeout{** the default language instead.}%
\else
\language=\csname l@#1\endcsname
\fi
#2}}
\providecommand{\BIBdecl}{\relax}
\BIBdecl

\bibitem{Thomas2016}
J.~{Thomas}, G.~{Loianno}, K.~{Daniilidis}, and V.~{Kumar}, ``Visual servoing
  of quadrotors for perching by hanging from cylindrical objects,'' \emph{IEEE
  Robotics and Automation Letters}, vol.~1, no.~1, pp. 57--64, 2016.

\bibitem{Mathe2015}
\BIBentryALTinterwordspacing
K.~Máthé and L.~Buşoniu, ``Vision and control for uavs: A survey of general
  methods and of inexpensive platforms for infrastructure inspection,''
  \emph{Sensors}, vol.~15, no.~7, pp. 14\,887--14\,916, 2015. [Online].
  Available: \url{https://www.mdpi.com/1424-8220/15/7/14887}
\BIBentrySTDinterwordspacing

\bibitem{BONYANKHAMSEH2018221}
\BIBentryALTinterwordspacing
H.~{Bonyan Khamseh}, F.~Janabi-Sharifi, and A.~Abdessameud, ``Aerial
  manipulation—a literature survey,'' \emph{Robotics and Autonomous Systems},
  vol. 107, pp. 221 -- 235, 2018. [Online]. Available:
  \url{http://www.sciencedirect.com/science/article/pii/S0921889017305535}
\BIBentrySTDinterwordspacing

\bibitem{lippiello2015hybrid}
V.~Lippiello, J.~Cacace, A.~Santamaria-Navarro, J.~Andrade-Cetto, M.~A.
  Trujillo, Y.~R.~R. Esteves, and A.~Viguria, ``Hybrid visual servoing with
  hierarchical task composition for aerial manipulation,'' \emph{IEEE Robotics
  and Automation Letters}, vol.~1, no.~1, pp. 259--266, 2015.

\bibitem{thomas2017autonomous}
J.~Thomas, J.~Welde, G.~Loianno, K.~Daniilidis, and V.~Kumar, ``Autonomous
  flight for detection, localization, and tracking of moving targets with a
  small quadrotor,'' \emph{IEEE Robotics and Automation Letters}, vol.~2,
  no.~3, pp. 1762--1769, 2017.

\bibitem{mueggler2014event}
E.~Mueggler, B.~Huber, and D.~Scaramuzza, ``Event-based, 6-dof pose tracking
  for high-speed maneuvers,'' in \emph{2014 IEEE/RSJ International Conference
  on Intelligent Robots and Systems}.\hskip 1em plus 0.5em minus 0.4em\relax
  IEEE, 2014, pp. 2761--2768.

\bibitem{Chuang2019tracking}
\BIBentryALTinterwordspacing
H.-M. Chuang, D.~He, and A.~Namiki, ``Autonomous target tracking of uav using
  high-speed visual feedback,'' \emph{Applied Sciences}, vol.~9, no.~21, 2019.
  [Online]. Available: \url{https://www.mdpi.com/2076-3417/9/21/4552}
\BIBentrySTDinterwordspacing

\bibitem{Falanga2019toofast}
D.~{Falanga}, S.~{Kim}, and D.~{Scaramuzza}, ``How fast is too fast? the role
  of perception latency in high-speed sense and avoid,'' \emph{IEEE Robotics
  and Automation Letters}, vol.~4, no.~2, pp. 1884--1891, 2019.

\bibitem{oliveira2013ground}
T.~Oliveira and P.~Encarna{\c{c}}{\~a}o, ``Ground target tracking control
  system for unmanned aerial vehicles,'' \emph{Journal of Intelligent \&
  Robotic Systems}, vol.~69, no.~1, pp. 373--387, 2013.

\bibitem{borowczyk2017autonomous}
A.~Borowczyk, D.-T. Nguyen, A.~Phu-Van~Nguyen, D.~Q. Nguyen, D.~Saussi{\'e},
  and J.~Le~Ny, ``Autonomous landing of a multirotor micro air vehicle on a
  high velocity ground vehicle,'' \emph{Ifac-Papersonline}, vol.~50, no.~1, pp.
  10\,488--10\,494, 2017.

\bibitem{zhang2018vision}
L.~Zhang, F.~Deng, J.~Chen, Y.~Bi, S.~K. Phang, X.~Chen, and B.~M. Chen,
  ``Vision-based target three-dimensional geolocation using unmanned aerial
  vehicles,'' \emph{IEEE Transactions on Industrial Electronics}, vol.~65,
  no.~10, pp. 8052--8061, 2018.

\bibitem{cheng2017autonomous}
H.~Cheng, L.~Lin, Z.~Zheng, Y.~Guan, and Z.~Liu, ``An autonomous vision-based
  target tracking system for rotorcraft unmanned aerial vehicles,'' in
  \emph{2017 IEEE/RSJ International Conference on Intelligent Robots and
  Systems (IROS)}.\hskip 1em plus 0.5em minus 0.4em\relax IEEE, 2017, pp.
  1732--1738.

\bibitem{horla2021tune}
D.~Horla, W.~Giernacki, T.~B{\'a}{\v{c}}a, V.~Spurny, and M.~Saska, ``Al-tune:
  A family of methods to effectively tune uav controllers in in-flight
  conditions,'' \emph{Journal of Intelligent \& Robotic Systems}, vol. 103,
  no.~1, pp. 1--16, 2021.

\bibitem{horla2021optimal}
D.~Horla, M.~Hamandi, W.~Giernacki, and A.~Franchi, ``Optimal tuning of the
  lateral-dynamics parameters for aerial vehicles with bounded lateral force,''
  \emph{IEEE Robotics and Automation Letters}, vol.~6, no.~2, pp. 3949--3955,
  2021.

\bibitem{Chehadeh2019}
\BIBentryALTinterwordspacing
M.~S. Chehadeh and I.~Boiko, ``{Design of rules for in-flight non-parametric
  tuning of PID controllers for unmanned aerial vehicles},'' \emph{Journal of
  the Franklin Institute}, vol. 356, no.~1, pp. 474--491, jan 2019. [Online].
  Available:
  \url{https://linkinghub.elsevier.com/retrieve/pii/S0016003218306604}
\BIBentrySTDinterwordspacing

\bibitem{ayyad2020real}
A.~{Ayyad}, M.~{Chehadeh}, M.~I. {Awad}, and Y.~{Zweiri}, ``Real-time system
  identification using deep learning for linear processes with application to
  unmanned aerial vehicles,'' \emph{IEEE Access}, vol.~8, pp.
  122\,539--122\,553, 2020.

\bibitem{Boiko2013book}
\BIBentryALTinterwordspacing
I.~Boiko, ``Modified relay feedback test (mrft) and tuning of pid
  controllers,'' in \emph{Non-parametric Tuning of PID Controllers: A Modified
  Relay-Feedback-Test Approach}.\hskip 1em plus 0.5em minus 0.4em\relax London:
  Springer London, 2013, pp. 25--79. [Online]. Available:
  \url{https://doi.org/10.1007/978-1-4471-4465-6_3}
\BIBentrySTDinterwordspacing

\bibitem{ayyad2021multirotors}
A.~Ayyad, M.~Chehadeh, P.~H. Silva, M.~Wahbah, O.~A. Hay, I.~Boiko, and
  Y.~Zweiri, ``Multirotors from takeoff to real-time full identification using
  the modified relay feedback test and deep neural networks,'' \emph{IEEE
  Transactions on Control Systems Technology}, 2021.

\bibitem{mbzirc2020}
\BIBentryALTinterwordspacing
``The mohamed bin zayed international robotics challenge 2020.'' [Online].
  Available: \url{https://www.mbzirc.com/challenge/2020}
\BIBentrySTDinterwordspacing

\bibitem{chaumette2016visual}
F.~Chaumette, S.~Hutchinson, and P.~Corke, ``Visual servoing,'' in
  \emph{Springer Handbook of Robotics}.\hskip 1em plus 0.5em minus 0.4em\relax
  Springer, 2016, pp. 841--866.

\bibitem{ALSHARMAN2018}
\BIBentryALTinterwordspacing
M.~K. Al-Sharman, B.~J. Emran, M.~A. Jaradat, H.~Najjaran, R.~Al-Husari, and
  Y.~Zweiri, ``Precision landing using an adaptive fuzzy multi-sensor data
  fusion architecture,'' \emph{Applied Soft Computing}, vol.~69, pp. 149--164,
  2018. [Online]. Available:
  \url{https://www.sciencedirect.com/science/article/pii/S1568494618302163}
\BIBentrySTDinterwordspacing

\bibitem{ALSHARMAN2020}
M.~K. Al-Sharman, Y.~Zweiri, M.~A.~K. Jaradat, R.~Al-Husari, D.~Gan, and L.~D.
  Seneviratne, ``Deep-learning-based neural network training for state
  estimation enhancement: Application to attitude estimation,'' \emph{IEEE
  Transactions on Instrumentation and Measurement}, vol.~69, no.~1, pp. 24--34,
  2020.

\bibitem{wahbah2021dynamic}
M.~Wahbah, M.~Chehadeh, and Y.~Zweiri, ``Dynamic based estimator for uavs with
  real-time identification using dnn and the modified relay feedback test,''
  2021.

\bibitem{Cheron2010}
\BIBentryALTinterwordspacing
C.~Cheron, A.~Dennis, V.~Semerjyan, and Y.~Chen, ``{A multifunctional HIL
  testbed for multirotor VTOL UAV actuator},'' in \emph{Proceedings of 2010
  IEEE/ASME International Conference on Mechatronic and Embedded Systems and
  Applications}.\hskip 1em plus 0.5em minus 0.4em\relax IEEE, jul 2010, pp.
  44--48. [Online]. Available:
  \url{http://ieeexplore.ieee.org/document/5552032/}
\BIBentrySTDinterwordspacing

\bibitem{ayyad2021tcst}
A.~Ayyad, M.~Chehadeh, P.~H. Silva, M.~Wahbah, O.~A. Hay, I.~Boiko, and
  Y.~Zweiri, ``Multirotors from takeoff to real-time full identification using
  the modified relay feedback test and deep neural networks,'' \emph{IEEE
  Transactions on Control Systems Technology}, pp. 1--17, 2021.

\bibitem{atherton1975}
D.~Atherton, \emph{Nonlinear Control Engineering: Describing Function Analysis
  and Design}.\hskip 1em plus 0.5em minus 0.4em\relax London: Van Nostrand
  Reinhold, 9 1975.

\bibitem{rehan2021optimal}
A.~Rehan, I.~Boiko, and Y.~Zweiri, ``Optimal non-parametric tuning of pid
  controllers based on classification of shapes of oscillations in modified
  relay feedback test,'' \emph{Journal of the Franklin Institute}, vol. 358,
  no.~2, pp. 1448--1474, 2021.

\bibitem{rohrer1965sensitivity}
R.~Rohrer and M.~Sobral, ``Sensitivity considerations in optimal system
  design,'' \emph{IEEE Transactions on Automatic Control}, vol.~10, no.~1, pp.
  43--48, 1965.

\bibitem{oussama_2021_video}
\BIBentryALTinterwordspacing
O.~AbdulHay, \emph{{Unified Identification and Tuning Approach Using DNN-MRFT
  For Visual Servoing Applications}}, 2021. [Online]. Available:
  \url{https://youtu.be/G69OldaoIKQ}
\BIBentrySTDinterwordspacing

\bibitem{Chaumette2018tracking}
B.~Penin, P.~R. Giordano, and F.~Chaumette, ``Vision-based reactive planning
  for aggressive target tracking while avoiding collisions and occlusions,''
  \emph{IEEE Robotics and Automation Letters}, vol.~3, no.~4, pp. 3725--3732,
  2018.

\bibitem{Dimitrova2020}
R.~S. Dimitrova, M.~Gehrig, D.~Brescianini, and D.~Scaramuzza, ``Towards
  low-latency high-bandwidth control of quadrotors using event cameras,'' in
  \emph{2020 IEEE International Conference on Robotics and Automation (ICRA)},
  2020, pp. 4294--4300.

\bibitem{Baca2019landing}
\BIBentryALTinterwordspacing
T.~Baca, P.~Stepan, V.~Spurny, D.~Hert, R.~Penicka, M.~Saska, J.~Thomas,
  G.~Loianno, and V.~Kumar, ``Autonomous landing on a moving vehicle with an
  unmanned aerial vehicle,'' \emph{Journal of Field Robotics}, vol.~36, no.~5,
  pp. 874--891, 2019. [Online]. Available:
  \url{https://onlinelibrary.wiley.com/doi/abs/10.1002/rob.21858}
\BIBentrySTDinterwordspacing

\bibitem{Falanga2019fast_too_fast}
D.~Falanga, S.~Kim, and D.~Scaramuzza, ``How fast is too fast? the role of
  perception latency in high-speed sense and avoid,'' \emph{IEEE Robotics and
  Automation Letters}, vol.~4, no.~2, pp. 1884--1891, 2019.

\bibitem{Vrba2022}
\BIBentryALTinterwordspacing
M.~Vrba, Y.~Stasinchuk, T.~Báča, V.~Spurný, M.~Petrlík, D.~Heřt,
  D.~Žaitlík, and M.~Saska, ``Autonomous capture of agile flying objects
  using uavs: The mbzirc 2020 challenge,'' \emph{Robotics and Autonomous
  Systems}, vol. 149, p. 103970, 2022. [Online]. Available:
  \url{https://www.sciencedirect.com/science/article/pii/S0921889021002396}
\BIBentrySTDinterwordspacing

\end{thebibliography}

\end{document}